\documentclass[letterpaper, 10pt, conference]{ieeeconf}  

\IEEEoverridecommandlockouts                              

\usepackage{graphics} 
\usepackage{epsfig} 
\usepackage{booktabs}
\usepackage{amsmath} 
\usepackage{amssymb}  
\usepackage{mathtools}

\usepackage{algorithm}
\usepackage{algpseudocode}

\usepackage{url}

\DeclareMathOperator*{\argmax}{arg\,max}
\DeclareMathOperator*{\argmin}{arg\,min}

\usepackage{tikz}

\newcommand\copyrighttext{%
  \footnotesize \textcopyright 2025 IEEE. Personal use of this material is permitted.
  Permission from IEEE must be obtained for all other uses, in any current or future
  media, including reprinting/republishing this material for advertising or promotional
  purposes, creating new collective works, for resale or redistribution to servers or
  lists, or reuse of any copyrighted component of this work in other works.
  }
\newcommand\copyrightnotice{%
\begin{tikzpicture}[remember picture,overlay]
\node[anchor=south,yshift=10pt] at (current page.south) {\fbox{\parbox{\dimexpr\textwidth-\fboxsep-\fboxrule\relax}{\copyrighttext}}};
\end{tikzpicture}%
}

\newcommand\acceptancetext{%
\centering
  \footnotesize Accepted for publication at the IEEE International Conference on Robotics and Automation (ICRA), 2025
  }
\newcommand\acceptancenotice{%
\begin{tikzpicture}[remember picture,overlay]
\node[anchor=north,yshift=-10pt] at (current page.north)
{\parbox{\dimexpr\textwidth-\fboxsep-\fboxrule\relax}
{\acceptancetext}};
\end{tikzpicture}%
} 

\title{\LARGE \bf
Visuo-Tactile Object Pose Estimation for a Multi-Finger Robot Hand with Low-Resolution In-Hand Tactile Sensing
}

\author{Lukas Mack$^{1,2}$, Felix Grüninger$^{3}$, Benjamin A. Richardson$^{4}$,\\ Regine Lendway$^{2}$, Katherine J. Kuchenbecker$^{4}$, and Joerg Stueckler$^{1,2}$
\thanks{*This work was supported by Cyber Valley, the Max Planck Society, and Hightech Agenda Bayern. 
This work was also supported by the Max Planck Institute for Intelligent Systems through Grassroots project GR1200 and by the Deutsche Forschungsgemeinschaft (DFG) project no.\ 466606396 (STU 771/1-1).}
\thanks{$^{1}$Lukas Mack and Joerg Stueckler are with the Intelligent Perception in Technical Systems Group, University of Augsburg, Augsburg, Germany
        {\tt\small firstname.lastname@uni-a.de}}%
\thanks{$^{2}$Lukas Mack, Regine Lendway and Joerg Stueckler are with the Embodied Vision Group, Max Planck Institute for Intelligent Systems, Tübingen, Germany
        {\tt\small firstname.lastname@tue.mpg.de}}%
\thanks{$^{3}$Felix Grüninger is with the Robotics ZWE, Max Planck Institute for Intelligent Systems, Tübingen, Germany
        {\tt\small grueninger@is.mpg.de}}%
\thanks{$^{4}$Benjamin A.\ Richardson and Katherine J.\ Kuchenbecker are with the Haptic Intelligence Department, Max Planck Institute for Intelligent Systems, Stuttgart, Germany
        {\tt\small \{richardson,kjk\}@is.mpg.de}}%
}

\begin{document}

\maketitle
\thispagestyle{empty}
\pagestyle{empty}
\acceptancenotice
\copyrightnotice 
\vspace{-\baselineskip} 

\begin{abstract}
Accurate 3D pose estimation of grasped objects is an important prerequisite for robots to perform assembly or in-hand manipulation tasks, but object occlusion by the robot's own hand greatly increases the difficulty of this perceptual task. 
Here, we propose that combining visual information and proprioception with binary, low-resolution tactile contact measurements from across the interior surface of an articulated robotic hand can mitigate this issue. 
The visuo-tactile object-pose-estimation problem is formulated probabilistically in a factor graph. 
The pose of the object is optimized to align with the three kinds of measurements using a robust cost function to reduce the influence of visual or tactile outlier readings. 
The advantages of the proposed approach are first demonstrated in simulation: a custom 15-DoF robot hand with one binary tactile sensor per link grasps 17 YCB objects while observed by an RGB-D camera. 
This low-resolution in-hand tactile sensing significantly improves object-pose estimates under high occlusion and also high visual noise. 
We also show these benefits through grasping tests with a preliminary real version of our tactile hand, obtaining reasonable visuo-tactile estimates of object pose at approximately 13.3\,Hz on average.
\end{abstract}

\section{Introduction}
Estimating the position and orientation (i.e., pose) of an object in 3D is a crucial capability for robotic manipulation of rigid everyday objects.
Many optimization- or learning-based control methods require an accurate object pose estimate to perform well in manipulation tasks such as grasping or in-hand reorientation~\cite{du2021_robotgraspingreview,handa2023_dextreme}.
The challenge of object-pose estimation from RGB and depth images has been studied intensively in the computer vision community in recent years.
There exist several highly capable deep learning-based approaches with or without object-model assumptions that process images and videos of scenes with a varying number of objects~\cite{he2021_ffb6d,wang2021_gdr,wen2023_foundationpose}.
However, when grasping an object, a robot end-effector often partially or fully occludes the object in the camera frame, diminishing the accuracy of visual pose-estimation methods.
At the same time, physically interacting with the object leads to additional information, such as geometric reasoning or tactile sensor measurements, that could be incorporated into the pose estimate.

While the most common robot end-effector for object manipulation is the two-fingered parallel gripper,
recent advances in building humanoid robots have drawn attention to multi-finger robot hands that can perform more challenging types of object manipulation~\cite{allegrohand, shadowhand, shaw2023leaphand, xu2023dexterous}. 
Robot hands with more fingers are generally capable of performing tasks more dexterously, including in-hand manipulation. 
However, these more complex hand morphologies and manipulation strategies naturally lead to much higher occlusion than the simpler tasks performed by parallel grippers.

\begin{figure}[t]
\centering
\includegraphics[width=0.99\linewidth]
{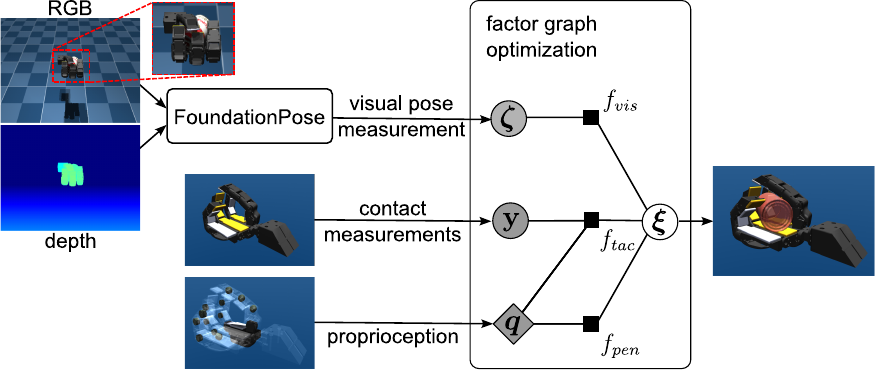}
\caption{Overview of our approach. We estimate the 3D pose of an object using factor graph optimization from visual pose, contact measurements, and proprioception. The visual pose measurement~$\boldsymbol{\zeta}\in SE(3)$ is obtained by a deep learning-based approach (FoundationPose~\cite{wen2023_foundationpose}) and is used in a visual factor~$f_{\mathit{vis}}$. Intersection constraints between the object and the hand in its current pose $\mathbf{q}$ are considered by a penetration factor~$f_{\mathit{pen}}$. Binary contact measurements~$\mathbf{y} \in \mathbb{B}^K$ are obtained per link from~$K$ rectangular sensor pads and are used in a tactile factor~$f_{\mathit{tac}}$ that also incorporates $\mathbf{q}$.}
\label{fig:teaser}
\end{figure}

In this work, we study the scenario of 3D object-pose estimation from visuo-tactile data obtained from a multi-finger robot hand equipped with low-resolution tactile contact sensing across the interior surface of the hand (palm and fingers).  
Specifically, we evaluate the approach for our custom-designed four-finger, 15-degree-of-freedom (DoF) robot hand (ISyHand v2) in simulation and with a preliminary real version of the hand. 
The inner surfaces of the palm and finger links are fully sensorized with 16 soft tactile sensor pads whose resistances change during contact.
Our visuo-tactile pose estimation method fuses vision-based pose estimates~\cite{wen2023_foundationpose} with tactile measurements and geometric constraints in a factor-graph-based optimization framework (see Fig.~\ref{fig:teaser}).
To determine expected tactile contact locations and geometric plausibility constraints with the hand, we model the object shape using a signed distance function (SDF), assuming that the shape is known, e.g., from a CAD model. 

We analyze our approach for tracking household objects inside the hand while grasping in simulation and in real-world experiments.
Our results show that the tactile sensor readings can improve the robustness of the object tracking in scenarios with high object occlusion or visual image noise.

In summary our contributions are:
\begin{itemize}
        \item We combine vision-based pose estimation with tactile measurements and geometric plausibility constraints for object-pose estimation in a multi-finger robot hand using factor graph optimization. 
        \item Our framework supports low-resolution tactile sensor pads covering the inner surfaces of palm and finger links without explicitly sensing contact location within each pad.
        \item Our visuo-tactile approach outperforms purely vision-based object tracking under high occlusion and noise. 
        \item We achieve real-time tracking with estimation at approximately 13.3\,Hz on average for our real robot setup.
\end{itemize}

\section{Related Work}
\subsubsection*{Vision-based pose estimation}
In recent years, several deep learning-based approaches have been proposed to estimate 3D object pose directly from RGB or RGB-D images~\cite{xiang2018_posecnn,he2021_ffb6d,wang2021_gdr,wang2019_nocs}.
Refinement methods have been proposed to further improve the accuracy of these pose estimators \cite{ li2018_deepim,lipson2022_coupled}; they compare the actual image with a rendering of the object in the estimated pose to improve the pose estimate. 
The recently proposed FoundationPose~\cite{wen2023_foundationpose} method combines neural pose hypothesis ranking for initial pose estimation with a transformer-based pose-refinement step. The model is trained on a large-scale synthetic dataset and achieves state-of-the-art results and tracking at high image rates. 
Yet, for the high levels of object occlusion that can occur during grasping, we have found that visual tracking alone can yield inaccurate results or fail to track the object.

\subsubsection*{Tactile-based pose estimation}
Several approaches have been proposed that use tactile contact measurements to estimate object pose.
The early approach in~\cite{koval2025_tacposeplanarpush} uses particle filters to iteratively determine a probability distribution over possible object poses during object pushing.
More recently, a particle filter estimates the 3D object pose from contact interactions~\cite{rostel2022_tacestimator}; using differentiable particle filters, this method learns neural proposal-distribution and particle-weighting functions in simulation which are transferred to real-world interactions by domain randomization.
Lin et al.~\cite{lin2023_tactileekf} use tactile sensor arrays with high spatial resolution and estimate the object pose during grasps in an extended Kalman filter~(EKF) framework.
In recent years, the availability of high-resolution vision-based tactile sensors has spurred new developments in tactile object-pose estimation~\cite{sodhi2022_patchgraph,kelestemur2022_tacpose,bauza2023_tac2pose,caddeo2023_collawaretacpose}.
In~\cite{caddeo2023_collawaretacpose}, for instance, the object pose is optimized by aligning deep learning-based predictions of the tactile images with measurements for multiple pose hypotheses and rejecting poses that penetrate the hand.

\subsubsection*{Object-pose estimation from vision and proprioception}
Rather than using tactile sensing, Liang et al.~\cite{liang2020_inhandposeparallelsim} employ proprioception for pose tracking and use vision-based pose estimation to initialize the tracker. 
Subsequently, the object is tracked using a physics simulation as a dynamics model, i.e., based on proprioception and geometric constraints.

\subsubsection*{Visuo-tactile pose estimation}
In this paper, we investigate fusing the complementary modalities of vision, tactile contact sensing, and proprioception for object-pose estimation.
In~\cite{bimbo2012_vistac}, vision is used to estimate an initial pose of the object using keypoints. Then, force-torque sensors in the fingertips are used to measure contact location and torque applied to the object and align the contact measurements with the 3D object point cloud.
Izatt et al.~\cite{izatt2017_vistac} devise measurement models for depth images of an RGB-D camera and GelSight tactile fingertip sensors for probabilistic filtering of the object pose. 
More recently,~\cite{zhao2023_fingerslam} combine a depth camera with a GelSight fingertip tactile depth sensor to track object pose and shape in a factor graph-based SLAM approach.
Dikhale et al.~\cite{dikhale2022_vistac} propose a deep learning-based approach that predicts object pose from RGB-D images of a camera along with depth images derived from tactile sensor measurements. 
In~\cite{anzai2020_vistac}, vision-based tactile measurements are input with RGB images into a network that learns to weight the modalities for estimating the relative object motion between frames.
Wan et al.~\cite{wan2024_vint6d} estimate object pose during grasping using an RGB-D camera and dense tactile-sensor arrays with high resolution on the full inner surface of the hand.
They fuse measurements from the two modalities using a deep learning-based approach that is trained on synthetic data of a set of object instances.
Differently, our approach uses simple, soft sensor pads which provide only a single scalar measurement per link.
We fuse this information in a factor graph with a state-of-the-art visual object-pose estimator (FoundationPose~\cite{wen2023_foundationpose}) which in principle can also be applied to novel objects unseen during training if a CAD model is available.

\section{Method}
\label{sec:method}

We formulate our visuo-tactile object-pose-estimation problem probabilistically in a factor graph.
The object pose is inferred from visual and tactile measurements together with geometric constraints between the hand and object.
The visual measurement is a 3D pose estimate (3 DoF position and 3 DoF rotation) from an object-pose estimator.
Each tactile measurement is a binary contact detection on a rectangular sensor pad; during optimization, each contact is localized to an estimated contact point.
Geometric constraints, which are derived from proprioception (joint angles), penalize interpenetrations between object and hand.

\subsection{Factor Graphs for Visuo-Tactile Sensor Fusion}
Formally, a factor graph is defined as a bipartite graph with nodes representing random variables $X_i, i\in\{1,\ldots,N\}$ with domains $\Omega_i$ and factors $f_j: \Omega(\mathcal{X}_j) \rightarrow \mathbb{R}$ which are scalar functions over subsets $\mathcal{X}_j \subseteq \{X_1,\ldots,X_N\}$ of the random variables~\cite{bishop2007_prml,dellaert2017_factorgraphs}.
The factors model a factorization of the joint probability distribution over the random variables, i.e., $p( x_1, \ldots, x_N ) = \prod_j f_j ( \mathcal{X}_j )$.
In state-estimation problems, the random variables represent observations and latent states which need to be estimated.
The factors model probabilistic dependencies between these variables such as priors, measurement likelihoods or probabilistic motion models.
For state estimation, we condition on the observed variables $\mathcal{X}_z$ and infer the latent state variables $\mathcal{X}_s$ by optimizing
    $\mathcal{X}_s^\ast = \argmax_{\mathcal{X}_s} \prod_j f_j (\mathcal{X}_j)$,
where the variables in~$\mathcal{X}_z$ are set to their given values.
If the factors represent local conditional probability distributions of a directed probabilistic graphical model, this problem corresponds to maximum a posteriori (MAP) estimation. 

We model our state estimation and sensor-fusion problem in a directed probabilistic graphical model whose factor graph is depicted in Fig.~\ref{fig:teaser}.
The pose $\boldsymbol{\xi} \in SE(3)$ of the object is the latent state (i.e., $\mathcal{X}_s = \{ \boldsymbol{\xi} \}$) which we seek to infer from the observed visual pose $\boldsymbol{\zeta} \in SE(3)$; $K$ binary tactile contact measurements $\mathbf{y} \in \{0,1\}^K$, where each dimension corresponds to one sensor pad; and the configuration of the hand joints~$\mathbf{q}$, i.e., $\mathcal{X}_z = \{ \boldsymbol{\zeta}, \mathbf{y}, \mathbf{q} \}$.
We model noisy visual and tactile measurements, but we assume accurate knowledge of the hand configuration and model it deterministically. 
A uniform state-transition model is assumed to avoid introducing bias with a simplifying assumption such as constant pose or velocity.
The optimization is initialized once with the vision pose estimate. In all subsequent time steps, it uses the pose estimate from the previous step. 
Since we optimize in single time steps, the time index is omitted in the following.

\subsection{Visual Pose Measurement Model}
The visual object measurement is obtained from a deep learning-based 3D object-pose estimator (FoundationPose~\cite{wen2023_foundationpose}).
It yields the pose measurement~$\boldsymbol{\zeta} \in SE(3)$, which we compare with the current object pose estimate~$\boldsymbol{\xi}$ by extracting twist coordinates for the relative pose error,
        $\mathbf{r}_{\mathit{vis}}(\boldsymbol{\xi}, \boldsymbol{\zeta}) = \left(\log\left( \boldsymbol{\xi}^{-1}\boldsymbol{\zeta} \right)\right)^\vee$
where~$\log$ is the matrix logarithm, which maps each element in $SE(3)$ to its associated Lie algebra $\mathfrak{se}(3)$, and the operator $\vee$ extracts the 6D twist vector from the Lie algebra element.
To facilitate real-time operation, we perform an initial pose estimate with FoundationPose only once at the start of each run and use its pose-refinement component thereafter.

\begin{figure}[t]
\centering
\vspace*{1.2ex}
\includegraphics[width=0.4\linewidth]{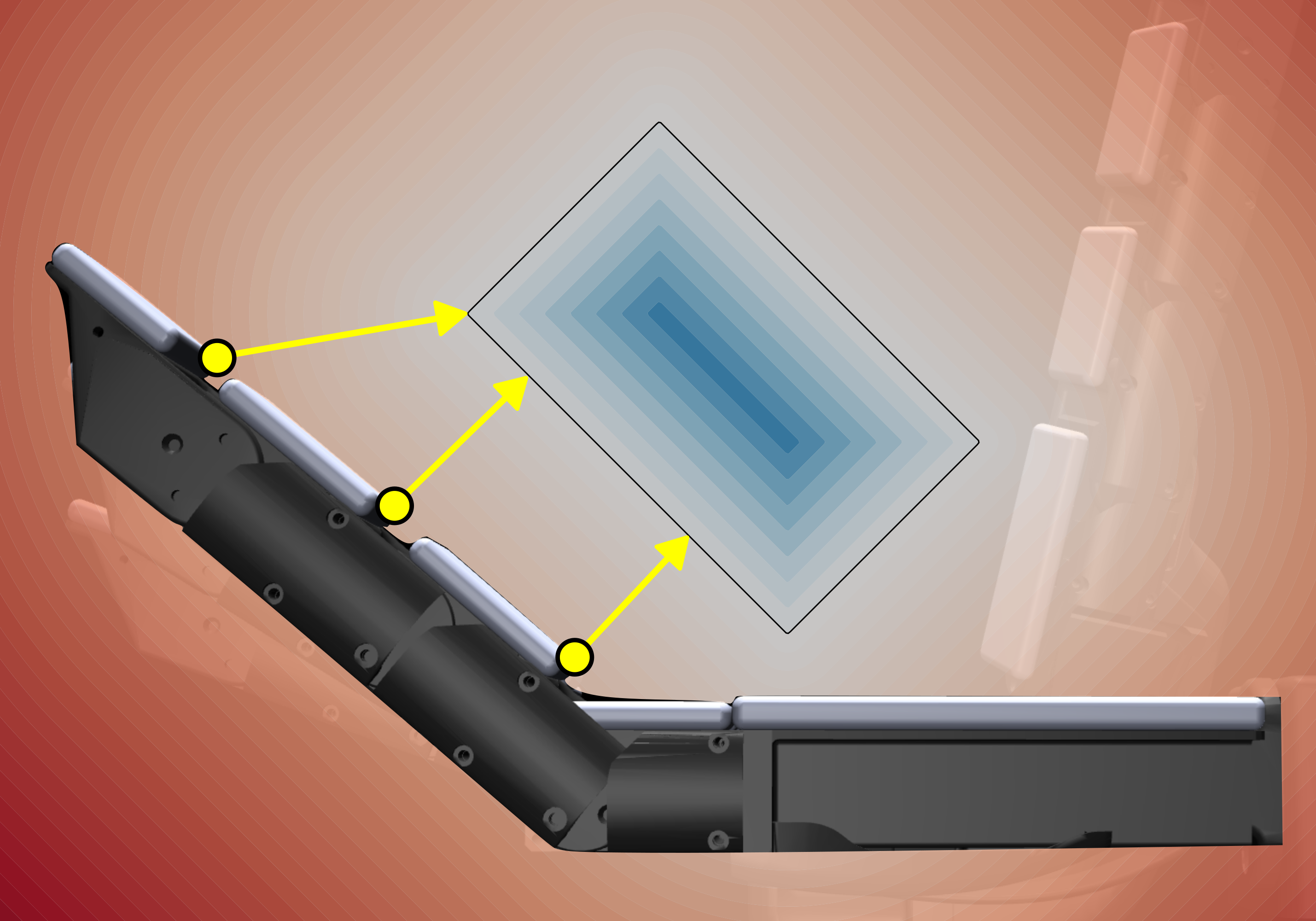}
\caption{Tactile residuals measure the signed distance of contact points on each sensor pad to the object if the object does not penetrate the pad.}
\label{fig:contactresidual}
\end{figure}

\subsection{Tactile Measurement Model}
\label{sec:tactilefactor}
For simplicity, we model our tactile sensors and the grasped object as rigid. When a tactile sensor pad detects contact with the object, the optimized object pose should have zero distance between the object surface and the sensor pad.
We evaluate this distance as the tactile measurement residual for each sensor pad~$k\in\{1,\dots,K\}$ using the closest point~$\mathbf{p}_{c,k} \in \mathbb{R}^3$ on the sensor pad to the object surface, where $K=16$ for our robot hand.
The residual 
\begin{equation}
        \mathbf{r}_{\mathit{tac},k}(\boldsymbol{\xi}, y_k,\mathbf{q})
        = y_k \cdot \max \left\{0, \phi_{\mathit{sdf}}\left(\boldsymbol{\xi}^{-1} \mathbf{p}_{c,k}\right) \right\}
\end{equation}
measures the signed distance of the closest point on the sensor pad to the object surface.
We use a signed distance function $\phi_{\mathit{sdf}}: \mathbb{R}^3 \rightarrow \mathbb{R}$ that is positive for points outside and negative for penetration into the object.
For each pad where contact is detected, this residual causes the optimization to pull the object surface onto the contact point.

The tactile sensor pads in our robot hand are shaped as cuboids and measure only one scalar value for pressure across each full pad.
The contact point $\mathbf{p}_{c,k} \in \mathbb{R}^3$ of the object on the sensor pad is not directly measured in the event of contact.
To determine the approximate contact point, we cover the sensor pad surface with a discrete rectangular grid of~$M$ points with equidistant spacing of 1\,mm.
The round sensor pad edges with a radius of 2\,mm are excluded from the planar grid to avoid implausible contact points.
The contact point is approximated by the grid point with the minimal SDF value (see Fig.~\ref{fig:contactresidual}). 
We precompute the object SDF from its mesh and discretize it into a 3D grid with resolution $128^3$.
Subvoxel-accurate SDF and gradient values are calculated using trilinear interpolation for a batch of contact points using the efficient CUDA kernel implementation\footnote{\url{https://github.com/EmbodiedVision/ev-sdf-utils}} of DiffSDFSim~\cite{strecke2021_diffsdfim}.
For points outside the 3D grid, the SDF value of the bounding box is approximately added to the SDF value of the closest point on the volume boundary.

\subsection{Non-Penetration Prior}
Given our assumption of rigidity, it is physically implausible that the object penetrates into any part of the hand.
Hence, we add a non-penetration prior for each sensor pad, which acts through a residual 
\begin{equation}
        \mathbf{r}_{\mathit{pen},k}(\boldsymbol{\xi},\mathbf{q})
        = \min\left\{ 0, \phi_{\mathit{sdf}}\left(\boldsymbol{\xi}^{-1} \mathbf{p}_{c,k}\right) \right\}
\end{equation}
that discourages negative SDF values in the object shape for the point of the object mesh closest to the sensor pad.
This point is readily determined like the above contact point as the point with minimal SDF value for each sensor pad.

\subsection{Non-Linear Optimization and Robust Cost Functions}

We assume the factors model normal distributions in the residuals, i.e., $f_j( \mathcal{X}_j ) = \mathcal{N}( \mathbf{r}_j( \mathcal{X}_{j,s}, \mathcal{X}_{j,z} ) \mid \mathbf{0}, \mathbf{\Sigma}_j )$, where $\mathbf{r}_j$ is a residual function, $\mathcal{X}_{j,s}$ and $\mathcal{X}_{j,z}$ are the latent state and observed variables in $\mathcal{X}_{j}$, respectively, and $\mathbf{\Sigma}_j$ is a covariance matrix.
Then, the MAP estimation problem becomes a non-linear least squares (NLS) problem by optimizing the negative logarithm of the posterior
\begin{equation}
\label{equ:NLS}
    \mathcal{X}_{s}^\ast = \argmin_{\mathcal{X}_{s}}  \frac{1}{2} \sum_j \lVert \boldsymbol{r}_j(\mathcal{X}_{j,s}, \mathcal{X}_{j,z}) \rVert^2_{\mathbf{\Sigma}_j},
\end{equation}
where~$\lVert \cdot \rVert^2_{\mathbf{\Sigma}}$ is a Mahalanobis distance with covariance $\mathbf{\Sigma}$.
This problem can be solved efficiently using second-order Gauss-Newton techniques.
To become robust against outlier measurements that break the assumption of the residuals being normally distributed, we use a sub-quadratic loss function $\rho(\cdot)$ in the optimization problem, as follows:
\begin{equation}
\label{equ:robustNLS}
    \mathcal{X}_{s}^\ast = \argmin_{\mathcal{X}_{s}}  \frac{1}{2} \sum_j \rho \left( w_j \lVert \boldsymbol{r}_j(\mathcal{X}_{j,s}, \mathcal{X}_{j,z}) \rVert^2 \right)
\end{equation}
with weights $w_j$ which correspond to an isotropic covariance.

We use the Welsch loss $\rho(e; \theta)= \theta-\theta\exp \left(\frac{-e}{\theta} \right)$ with hyperparameter~$\theta$ to reduce the influence of outliers in the visual and tactile measurements.
In the visual pose estimate, such outliers typically occur during high object occlusion.
On the real robot hand, the tactile sensors sporadically detect contact when none exists due to our currently simple signal-processing scheme.
We implement the optimization using Theseus~\cite{theseus} and its Levenberg-Marquardt~(LM) solver.

\section{Experiments}
We evaluate our approach in simulation and the real world by grasping household objects from the YCB dataset~\cite{Calli17-IJRR-YCB} with our custom ISyHand robot hand.
Observations are obtained from 16 tactile sensor pads that cover the inner-hand surface of the ISyHand and from a single RGB-D camera placed in an observing perspective. The general setup mimics a robot looking at its hand while grasping an object.
We compare our visuo-tactile approach with the baseline of FoundationPose~\cite{wen2023_foundationpose}.
Table~\ref{tab:hyperparam} reports the hyperparameter settings for both simulated and real experiments.

\begin{table}[b] 
\caption{Parameter settings.}
\vspace*{-3ex}
\label{tab:hyperparam}
\begin{center}
\begin{tabular}{lcc}
\toprule
 & \textbf{Simulation} & \textbf{Real World} \\
\midrule
Maximum LM solver iterations & 5 & 2 \\
LM damping & 0.02 & 0.02 \\
$\log(\theta_{vis})$ & $-6.5$ & $-5$ \\
$\log(\theta_{tac})$ & $0$ & $-7.5$ \\
\bottomrule
\end{tabular}
\end{center}
\end{table}

\subsection{ISyHand and Tactile Sensor Hardware}
The ISyHand (pronounced ``easy hand'') is a custom four-finger, 15-DoF fully articulated robot hand. Each finger has 3 DoF with parallel revolute joints, and the thumb has 4 DoF; the palm has 2 DoF but remained static during our experiments. Each joint is actuated with a Dynamixel XL330-M288-T motor, and the connecting links are customized 3D-printed components. Each link of the fingers and thumb is equipped with a single tactile sensor pad derived from those of Burns et al.~\cite{Burns22-FRAI-Endowing}.  As they are larger, the palm links each have two pads connected electrically in parallel to provide a single tactile input per link. The resistance of each pad changes when it is touched. 
Because the tactile sensor pads are handmade and have different sizes, they vary in their baseline resistance and sensitivity to touch. 
In general, normal contact pressure decreases a pad's resistance, whereas shear increases it. Since the baseline resistance can change after sustained contact, we calibrate this value before each experiment run. The resistance also temporarily increases beyond the initial baseline value when releasing contact, so we consider only negative changes from the baseline to detect contacts for pressure in the normal direction.
We compensate for the differing sensitivity across pads by normalizing each sensor scale to the resistance change measured when each pad is pressed with the same force of 6\,N.
Due to the simple thresholded contact detection, we evaluate our approach only for grasps that establish contact once in a run.
We also do not remove contacts caused by self-collisions of the hand.
While they are sufficient for proving the concept of our visuo-tactile approach to object-pose estimation, the preliminary design of these sensors and the associated signal-processing methods will be improved in future work.

\begin{figure}[tb]
    \centering
    \vspace{0.6ex}
    \includegraphics[width=0.99\linewidth]
    {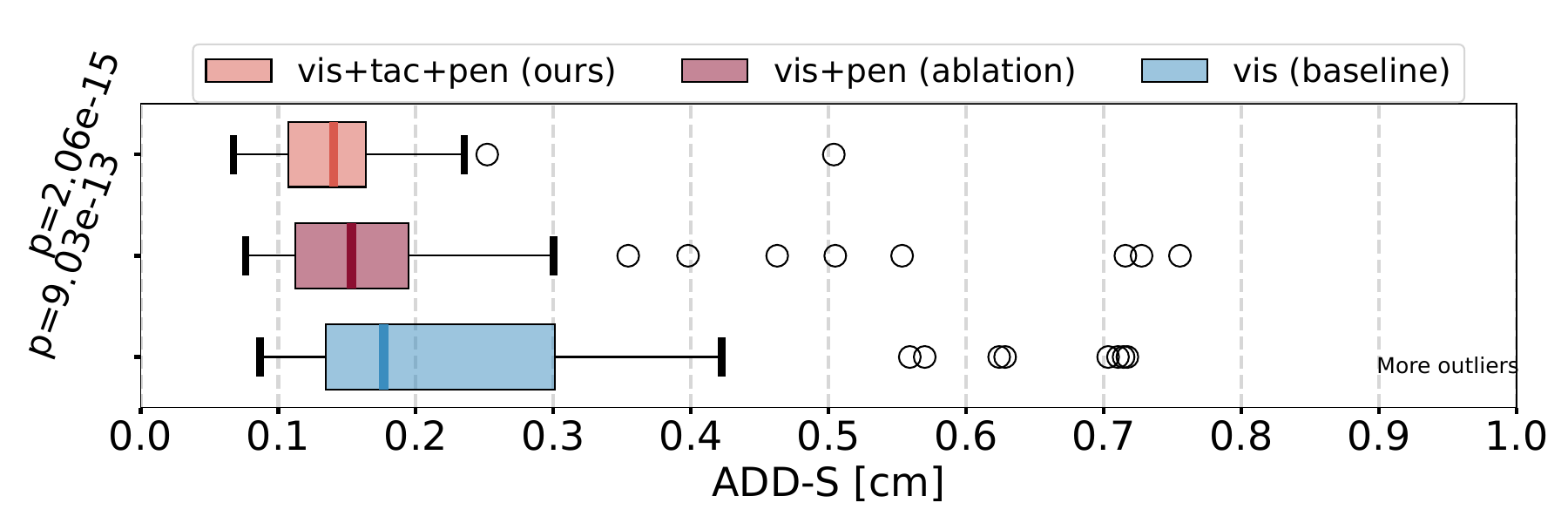}
\vspace*{-4ex}
    \caption{ADD-S statistics on the $D_{\mathit{vary}}$ dataset, comparing our visuo-tactile pose optimization against an ablation without tactile contacts and the vision baseline.
    Fusing vision and tactile information performs significantly better than both the ablation and the vision estimate (both $p<0.0001$).}
    \label{fig:nd_median_all}
\end{figure}

\begin{figure}[tb]
    \centering
    \vspace{0.6ex}
    \includegraphics[width=0.99\linewidth]
    {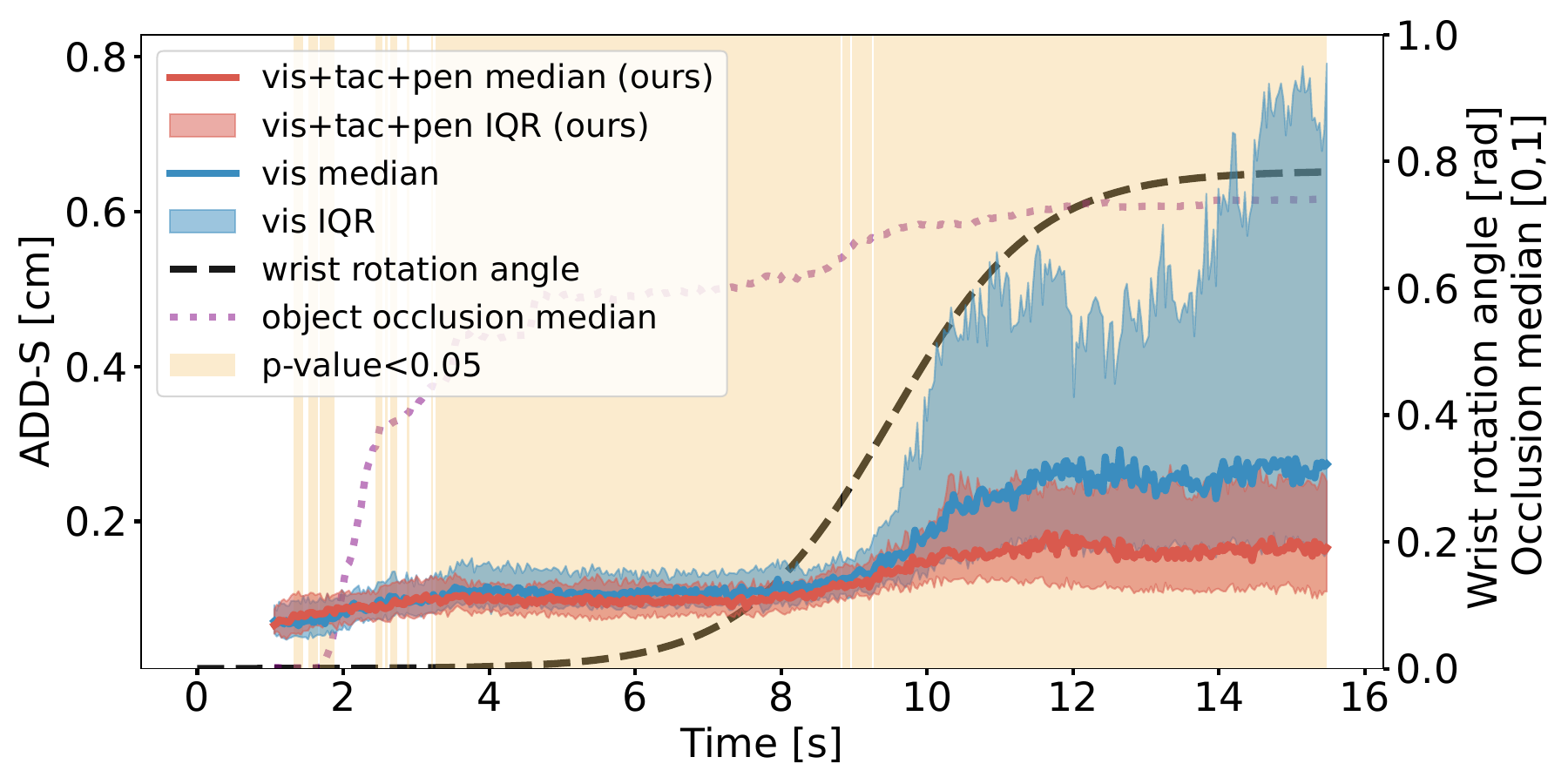}
\vspace*{-4ex}
    \caption{ADD-S statistics at each time step on the $D_{\mathit{vary}}$ dataset, comparing our visuo-tactile pose optimization against the vision baseline.  
    Fusing vision and tactile information significantly improves the accuracy of the pose estimate when the hand is rotated and the occlusion increases (after about 10\,s), and it provides similar accuracy before the rotation.}
    \label{fig:nd_median_time}
\end{figure}

\begin{figure}[tb]
    \centering
    \vspace{0.6ex}
    \includegraphics[width=0.95\linewidth]
    {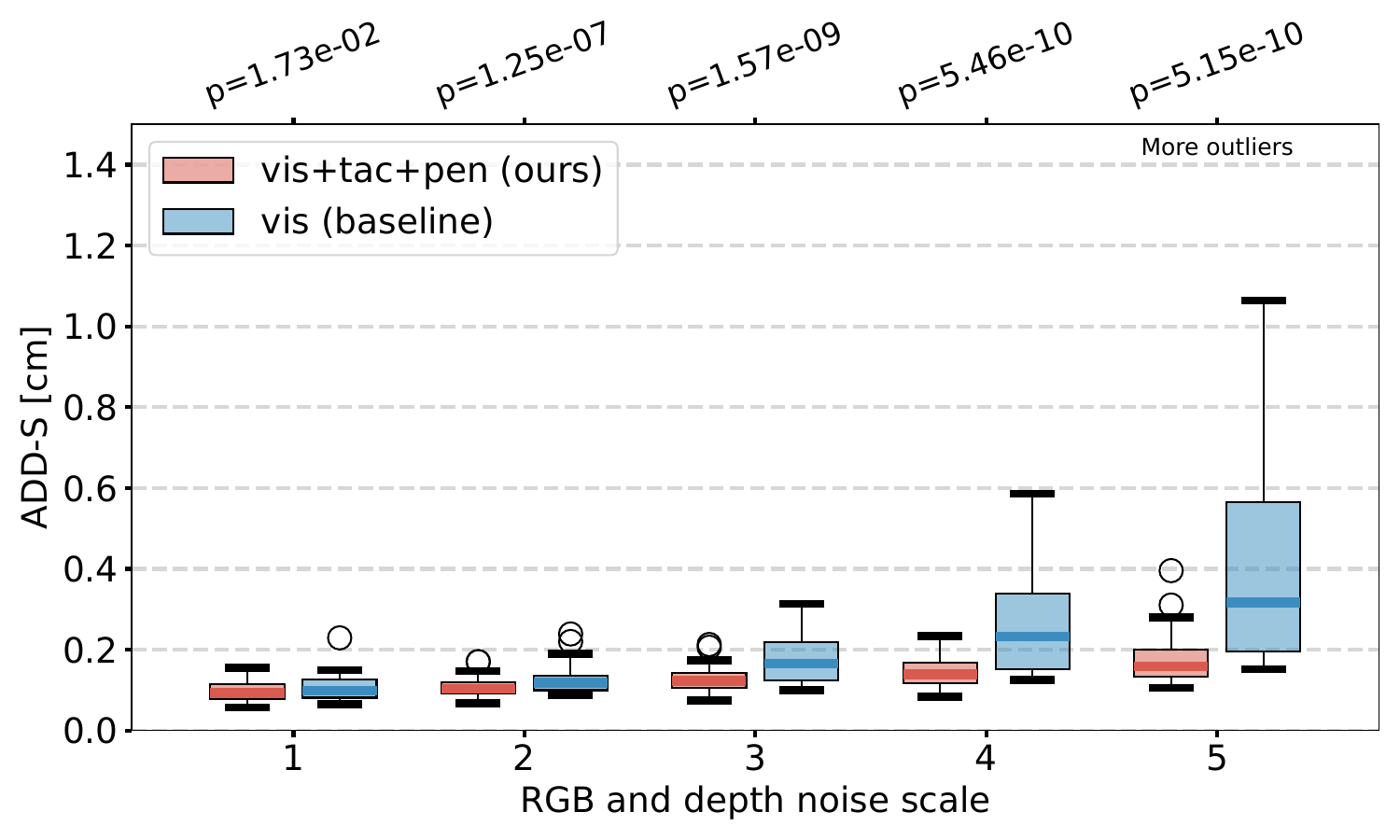}
\vspace*{-1ex}
    \caption{ADD-S statistics on $D_{\mathit{vary}}$ during the first 4.5\,s of the object grasps for five noise scales.
    Our approach of including tactile and penetration information improves accuracy significantly for all scales ($p<0.05$).}
    \label{fig:nd_median_noise}
\end{figure}

\begin{figure}[tb]
    \centering
    \vspace{0.6ex}
    \includegraphics[width=0.99\linewidth]
    {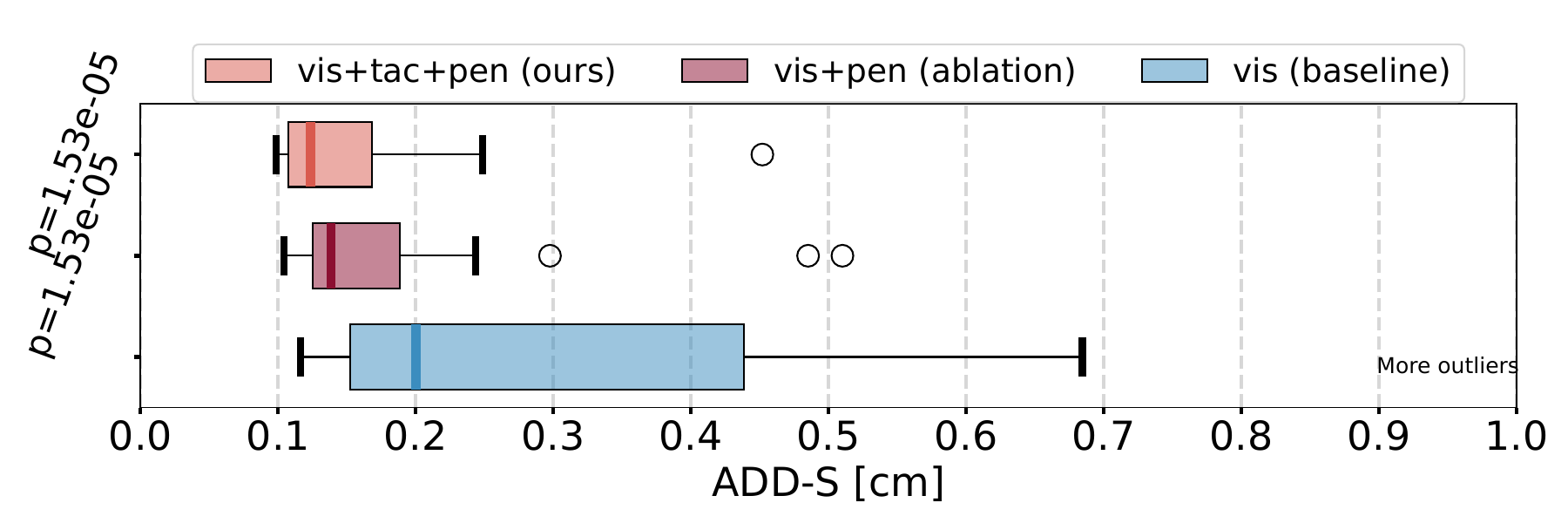}
\vspace*{-4ex}
    \caption{ADD-S statistics on the $D_{\mathit{occ}}$ dataset comparing our visuo-tactile pose optimization against an ablation without tactile contacts and the vision baseline. Fusing vision and tactile information significantly improves accuracy ($p<0.0001$). Note that here we average over the 5 runs per object to satisfy the statistical test's assumption of independent samples.}
    \label{fig:od_median_all}
\end{figure}

\begin{figure}[tb]
    \centering
    \vspace{0.6ex}
    \includegraphics[width=0.99\linewidth]
    {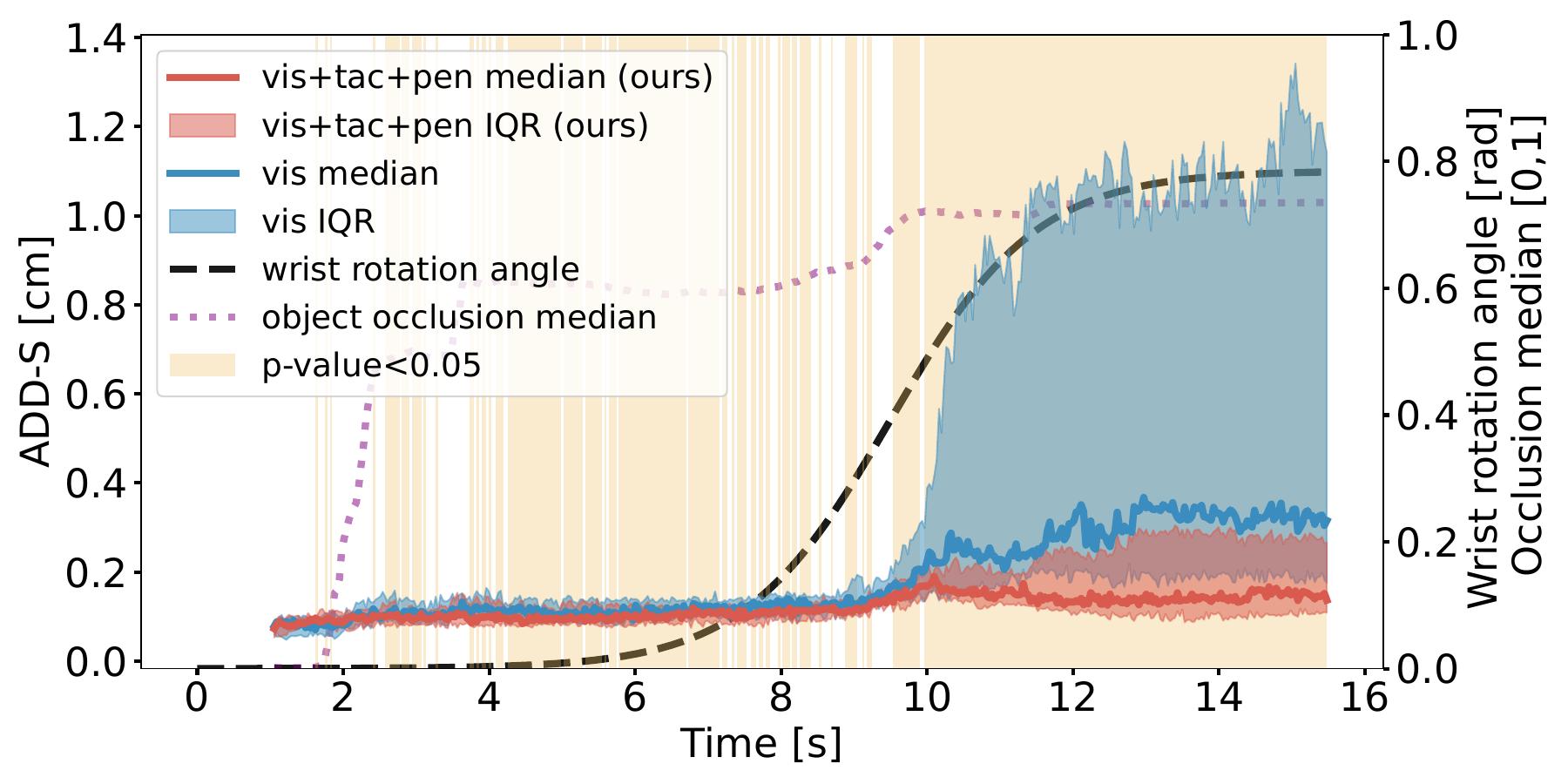}
\vspace*{-4ex}
    \caption{ADD-S statistics at each time step on the $D_{\mathit{occ}}$ dataset, with runs averaged by object. Comparing our approach against the vision baseline shows that 
    fusing vision and tactile information significantly improves pose accuracy when the hand is rotated and occlusion is high (after about 11\,s), and it provides similar accuracy before the rotation.} 
    \label{fig:od_median_time}
\end{figure}

\subsection{Simulation Experiment Setup}

We additionally develop a model of the ISyHand for use in simulation. To avoid modeling the complex mechanics of the tactile sensors, they are simulated as rigid elements that output binary contact signals for this work.
Note that the simulator provides measurements only for contacts between the object and sensor pads, filtering out self-collisions of the hand.
We simulate an RGB-D camera with intrinsics similar to the Intel RealSense D435 camera.
The camera captures images of the hand and the object by looking at the palm from a horizontal and vertical distance of 0.6\,m and 0.576\,m, respectively (see Fig.~\ref{fig:teaser}).
We simulate noise in the RGB and depth images by additive zero-mean Gaussian noise.
The depth noise standard deviation~$\sigma_{d}(d_{pix}) = 0.001063+0.0007278 \cdot d_{pix}+0.003949 \cdot d_{pix}^2$ is modelled pixel-wise as a quadratic function of the measured depth~$d_{pix} \in \mathbb{R}$~\cite{ahn2019_d435}.
To model RGB noise, the average standard deviation $\sigma_{rgb}=\sqrt{2.6}$ for a pixel was measured empirically with the Intel RealSense D435 camera in our real robot setup.
We use MuJoCo~\cite{todorov2012_mujoco} with a time step of 0.001\,s, elliptic friction cones, the implicitfast integrator, and default friction and solver parameters.
For our datasets we generate images at $1/0.033 \approx 30$\,Hz and tactile measurements at 100\,Hz.

\subsubsection*{Datasets}
We evaluate with 17 YCB objects~\cite{Calli17-IJRR-YCB} that the ISyHand can grasp from an open horizontal pose by closing its fingers and thumb.
We generated two datasets in which each object is grasped five times, resulting in $85$ sequences.
After the grasp, the hand is incrementally rotated backward by~$45^{\circ}$ to orient the fingers toward the camera and increase the occlusion of the object.
The first dataset~$D_{\mathit{vary}}$ samples the object's initial 3D orientation uniformly.
The object is placed at a default position at the center of the palm. 
An offset to the initial vertical position is sampled uniformly from zero to half the length of the object's shortest bounding-box edge.
Samples that penetrate the hand or do not yield a stable grasp are rejected.
The second dataset~$D_{\mathit{occ}}$ is designed to yield stable power grasps with high coverage of the object surface by the hand.
The initial object pose is chosen so that the object lies stably in the hand.
We add small perturbations to the object's rotation around the vertical axis by sampling an offset angle uniformly within $[-5^\circ,5^\circ]$.
In both datasets, we wait 1\,s at the beginning of the sequence so that the object is mostly static in the palm before grasping and starting the tracker.

\begin{figure*}[tbp!]
\vspace*{1.2ex}
\centering
\includegraphics[width=0.11\linewidth, trim={4.5cm 0.4cm 1.0cm 0}, clip]{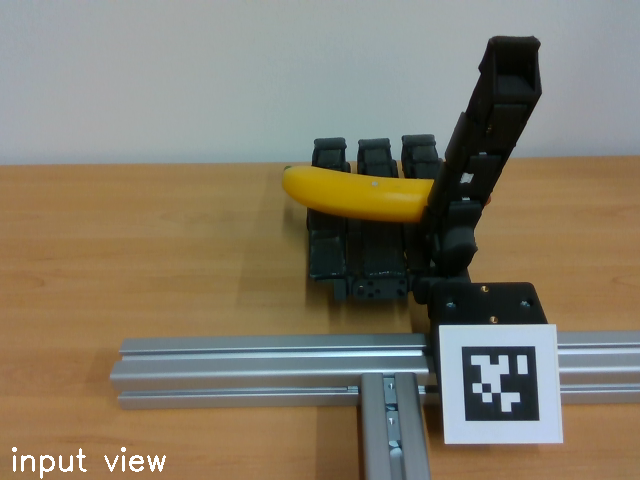}
\includegraphics[width=0.11\linewidth, trim={4.5cm 0.4cm 1.0cm 0}, clip]
{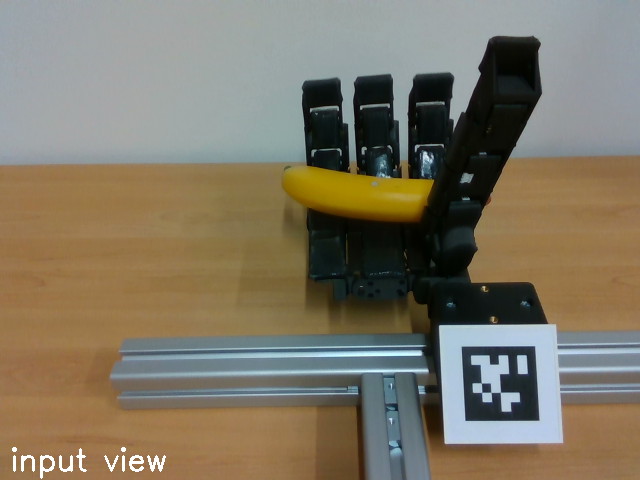}
\includegraphics[width=0.11\linewidth, trim={4.5cm 0.4cm 1.0cm 0}, clip]
{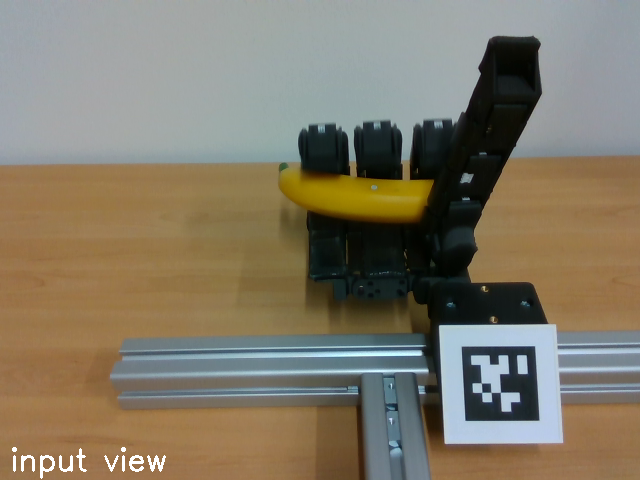}
\includegraphics[width=0.11\linewidth, trim={4.5cm 0.4cm 1.0cm 0}, clip]
{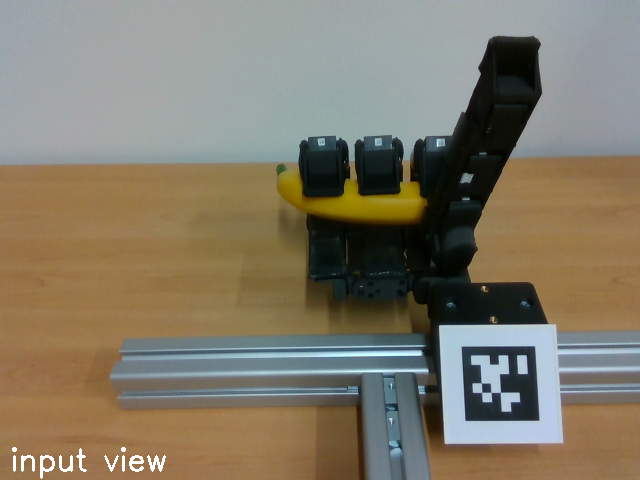} 
\hfill
\includegraphics[width=0.11\linewidth, trim={4.5cm 0.4cm 1.0cm 0}, clip]{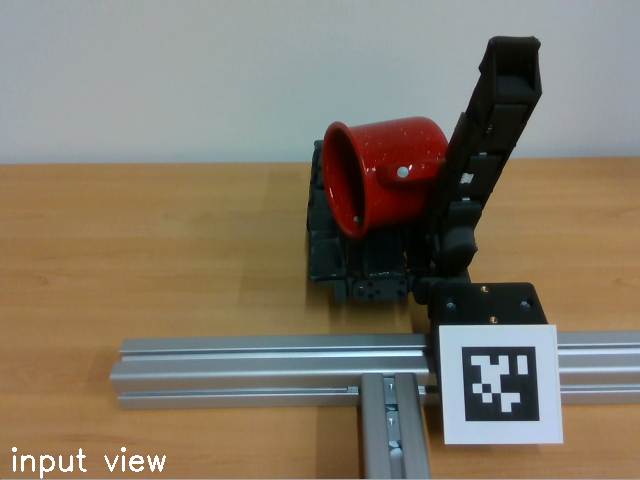}
\includegraphics[width=0.11\linewidth, trim={4.5cm 0.4cm 1.0cm 0}, clip]
{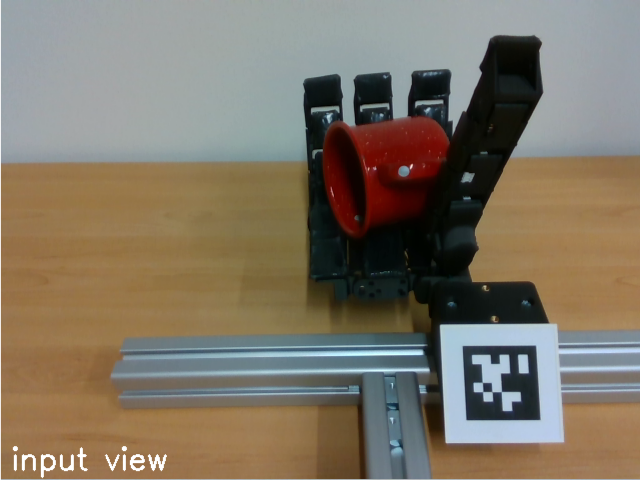}
\includegraphics[width=0.11\linewidth, trim={4.5cm 0.4cm 1.0cm 0}, clip]
{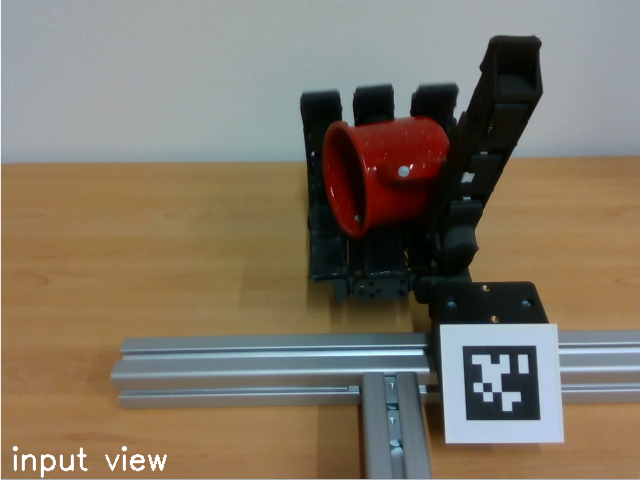}
\includegraphics[width=0.11\linewidth, trim={4.5cm 0.4cm 1.0cm 0}, clip]
{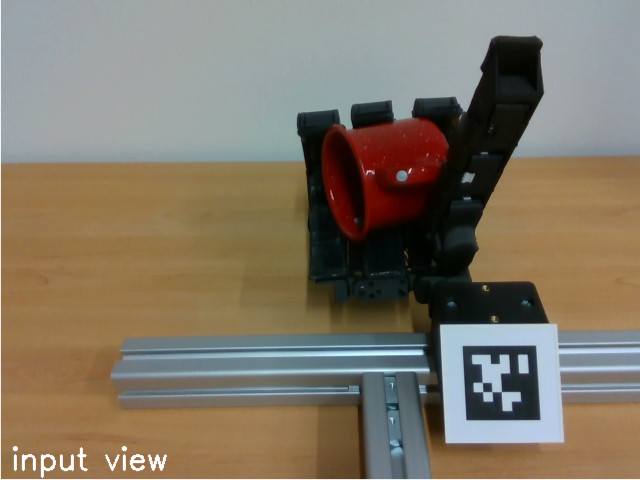}
\\
\vspace{0.5ex}
\includegraphics[width=0.11\linewidth, trim={6cm 1.1cm 0.0cm 0}, clip]{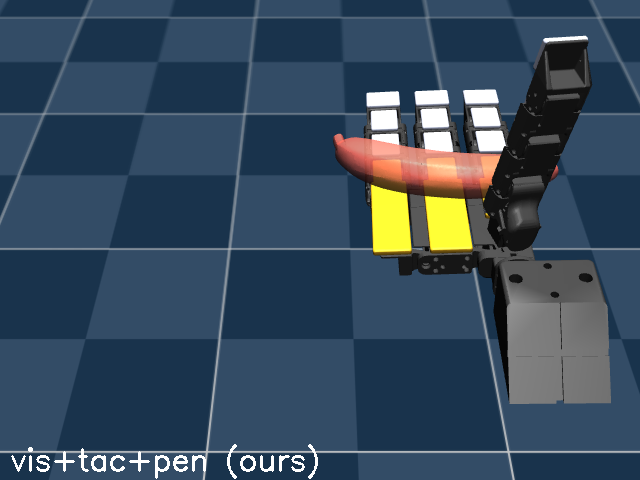}
\includegraphics[width=0.11\linewidth, trim={6cm 1.1cm 0.0cm 0}, clip]
{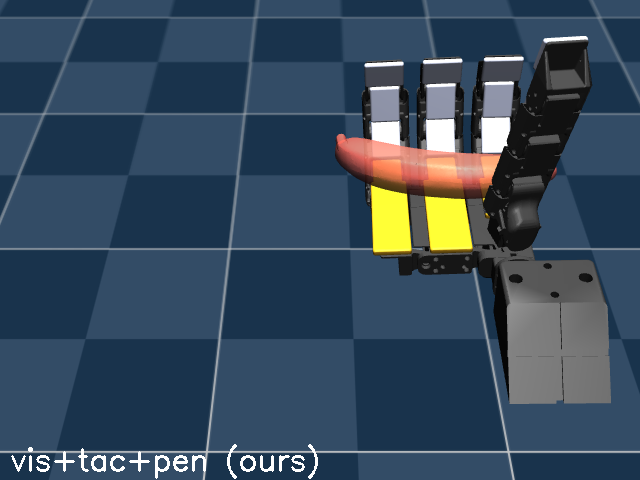}
\includegraphics[width=0.11\linewidth, trim={6cm 1.1cm 0.0cm 0}, clip]
{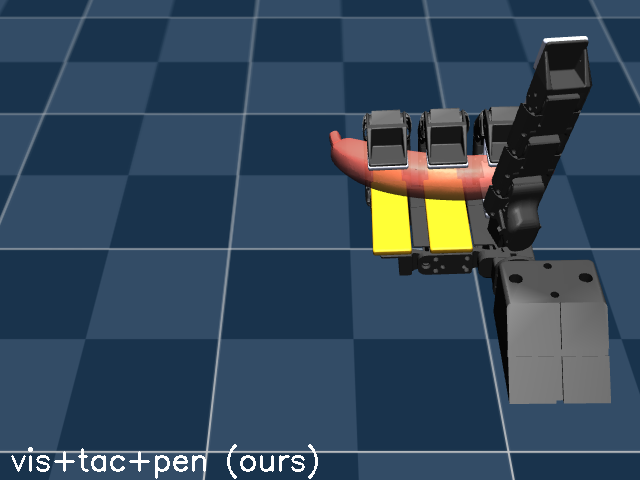}
\includegraphics[width=0.11\linewidth, trim={6cm 1.1cm 0.0cm 0}, clip]
{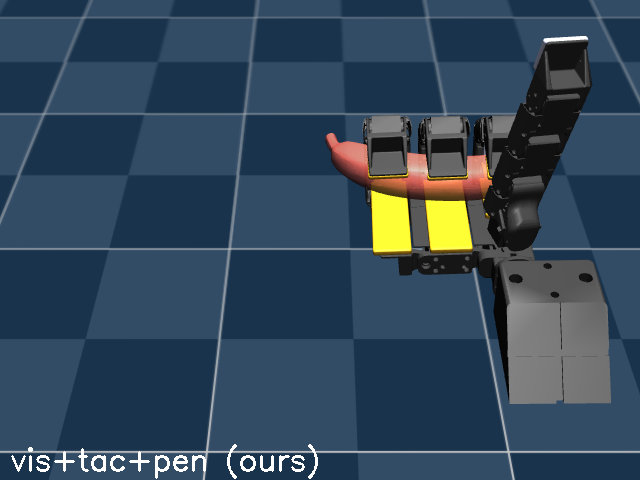}
\hfill
\includegraphics[width=0.11\linewidth, trim={6cm 1.1cm 0.0cm 0}, clip]{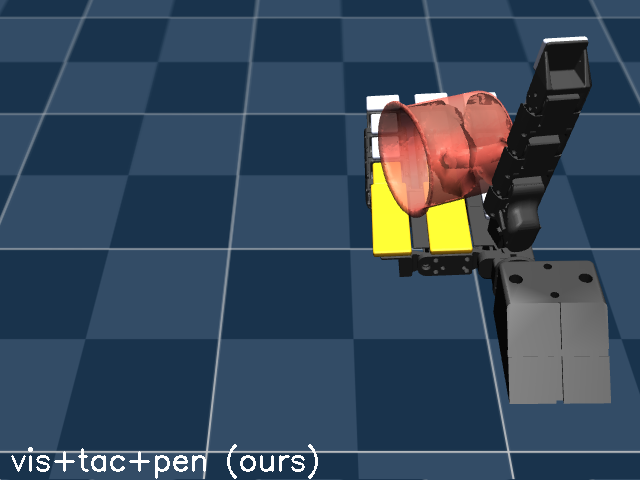}
\includegraphics[width=0.11\linewidth, trim={6cm 1.1cm 0.0cm 0}, clip]
{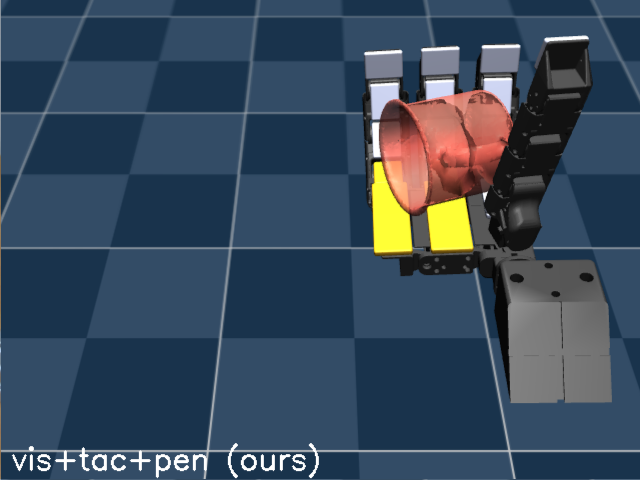}
\includegraphics[width=0.11\linewidth, trim={6cm 1.1cm 0.0cm 0}, clip]
{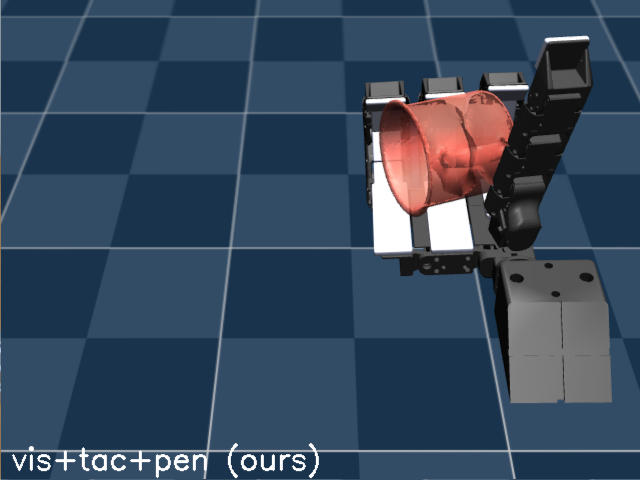}
\includegraphics[width=0.11\linewidth, trim={6cm 1.1cm 0.0cm 0}, clip]
{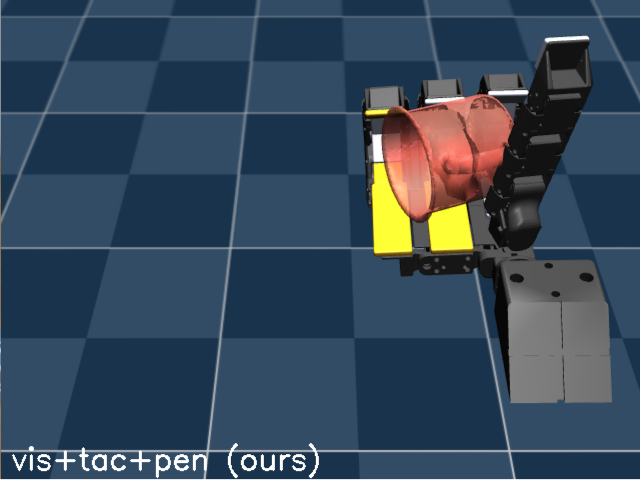}
\\
\vspace{0.5ex}
\includegraphics[width=0.11\linewidth, trim={6cm 1.1cm 0.0cm 0}, clip]{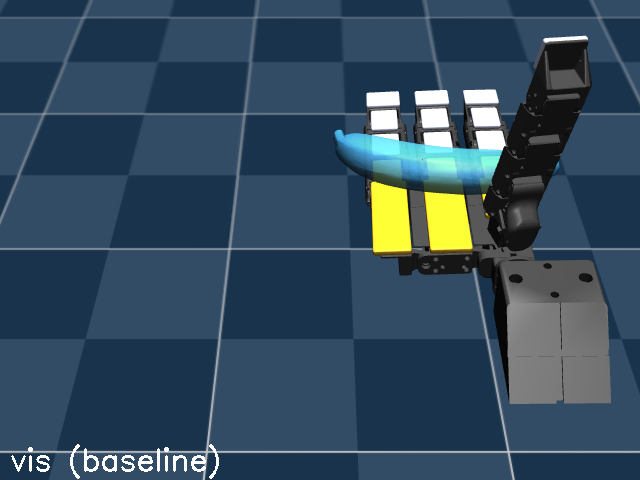}
\includegraphics[width=0.11\linewidth, trim={6cm 1.1cm 0.0cm 0}, clip]
{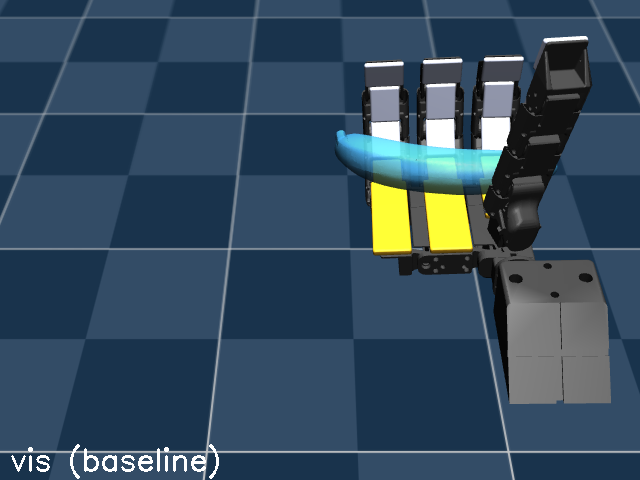}
\includegraphics[width=0.11\linewidth, trim={6cm 1.1cm 0.0cm 0}, clip]
{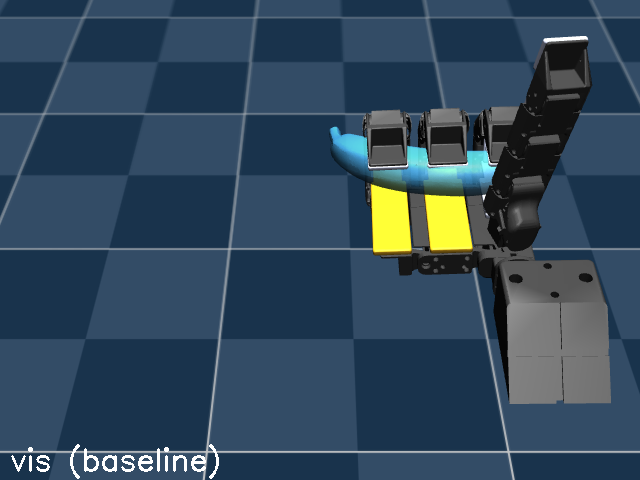}
\includegraphics[width=0.11\linewidth, trim={6cm 1.1cm 0.0cm 0}, clip]
{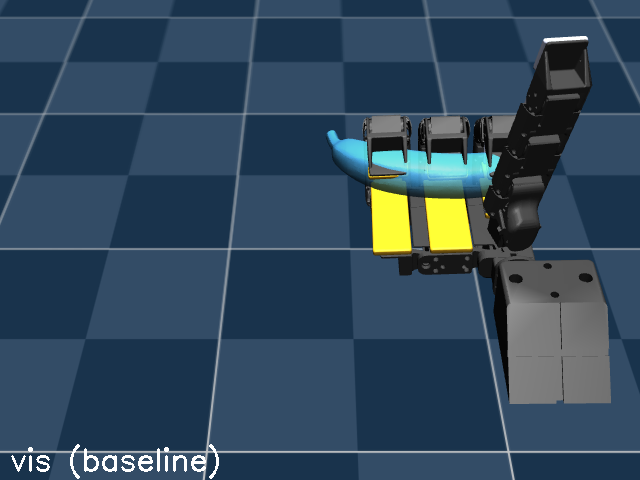}
\hfill
\includegraphics[width=0.11\linewidth, trim={6cm 1.1cm 0.0cm 0}, clip]{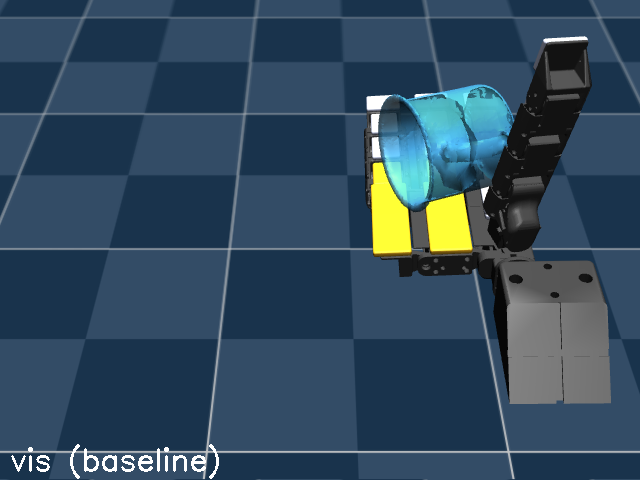}
\includegraphics[width=0.11\linewidth, trim={6cm 1.1cm 0.0cm 0}, clip]
{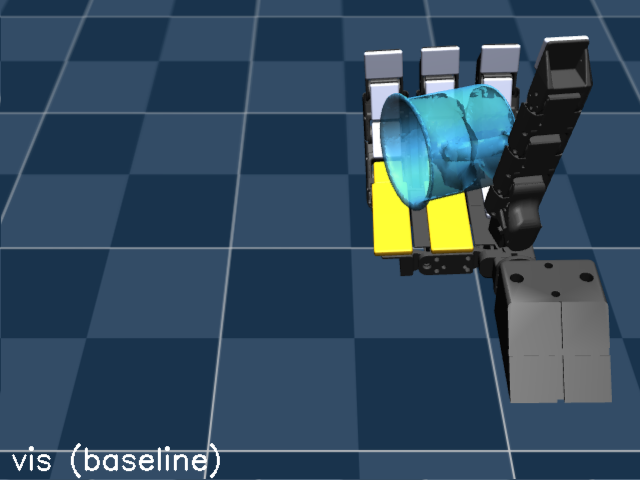}
\includegraphics[width=0.11\linewidth, trim={6cm 1.1cm 0.0cm 0}, clip]
{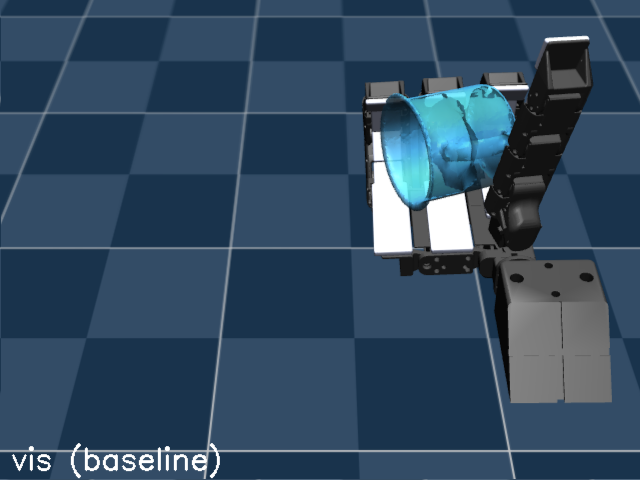}
\includegraphics[width=0.11\linewidth, trim={6cm 1.1cm 0.0cm 0}, clip]
{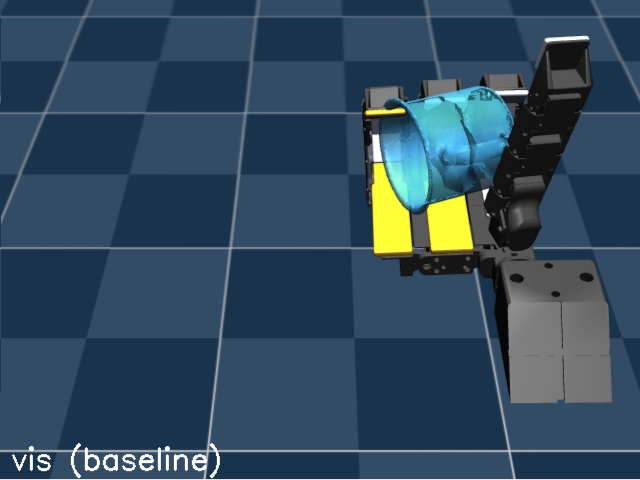}
\caption{Results with the real ISyHand. Left: Banana. Right: Mug. 
For each object: RGB image (top row), our visuo-tactile pose estimate (middle row), and the vision-only pose estimate (bottom row). Visuo-tactile pose estimation corrects for misalignments of the vision-based pose estimate and makes the output geometrically more plausible by considering contact measurements (yellow pads are detecting contact). Note interpenetration in the visual estimates.
}
\label{fig:realworldresults}
\end{figure*}

\subsection{Real-World Experiment Setup}

We place our real ISyHand in a similar horizontal posture as in the simulation experiments.
An Intel Realsense D435 RGB-D camera is mounted 0.31\,m behind and 0.32\,m above the hand and looks toward the wrist onto the hand.
The wrist pose is measured by an April tag marker averaged over the first 30 images. 
Images are captured at 30\,Hz and proprioceptive and tactile data at 50\,Hz, and we execute our optimization at 13.3\,Hz.
We evaluate real grasps for four YCB objects (banana, mug, mustard bottle, tomato soup can).

\subsection{Grasp Motion Control}
We design a heuristic grasp controller that closes the fingers and thumb by controlling the joints to follow an incremental position offset.
The joints close at different times: the inner joints close first, and the thumb starts closing after the fingers have closed.
In simulation, the joints are modelled as position servos with feedback parameter gains $(p_{prox}, p_{mid}, p_{dist})=(1.0, 0.7, 0.3)$ that are empirically tuned for each finger joint group (proximal, middle, distal).
For the real hand, we operate the Dynamixels in their position-current control mode.
The corresponding PID feedback gains $(p_{P}, p_{I}, p_{D})$ are set to $(400, 0, 400)$ for all finger joints and to $(1000,0,100)$ for the palm joints.

\subsection{Simulation Results}

In simulation, we know the ground-truth object pose and assess the accuracy of the pose estimates using the ADD-S measure~\cite{hinterstoisser2012_tless,xiang2018_posecnn}.
ADD-S is the average minimum distance of each object mesh vertex in the estimated pose to the mesh surface in the ground-truth pose and is invariant to shape symmetries.

\subsubsection*{Results on $D_{\mathit{vary}}$}
Fig.~\ref{fig:nd_median_all} shows evaluation results on the $D_{\mathit{vary}}$ dataset.
We observe that our visuo-tactile approach (vis+tac+pen) yields an overall better performance in median and IQR compared to the pure vision-based tracker (vis) that uses FoundationPose~\cite{wen2023_foundationpose}.
For statistical pairwise comparison of the reported ADD-S results, we apply the Wilcoxon signed-rank test between vis+tac+pen and vis ($p<0.0001$), confirming the significant improvement.
Furthermore, the inclusion of tactile measurements in our approach (vis+tac+pen) achieves significantly smaller errors compared to the ablation (vis+pen, $p<0.0001$).
From Fig.~\ref{fig:nd_median_time}, it can be seen that adding tactile and penetration factors achieves similar accuracy as vision-only tracking at the beginning of the sequences.
When the hand rotates and occlusion increases, the tactile and penetration factors improve the accuracy of the tracker significantly ($p<0.05$).
Videos of sample results can be found in the supplementary material.
In Fig.~\ref{fig:nd_median_noise} we evaluate the impact of various image noise levels on the tracking performance during the first 4.5\,s of each grasp, before the hand rotates.
For this experiment, we scale the standard deviation of the RGB and depth noise by various factors.
Tactile and penetration information significantly improves accuracy across all scales ($p<0.05$).

\subsubsection*{Results on $D_{\mathit{occ}}$}
Fig.~\ref{fig:od_median_all} compares the performance in the sequences with stronger occlusions when the hand is rotated.
Because the initial poses of each object are similar in these experiments, we don't assume sample independence and instead average the ADD-S values over the five runs per object.
Our approach (vis+tac+pen) demonstrates a clear improvement over the vision-only (vis) pose tracking ($p<0.0001$).
It also outperforms the ablation variant of our method without the tactile factors (vis+pen, $p<0.0001$).
Fig.~\ref{fig:od_median_time} shows the accuracy of the methods over time for~$D_{\mathit{occ}}$.
While including tactile and penetration information provides similar performance as vision-only tracking before the hand is rotated, both tactile and penetration information significantly improve accuracy for higher occlusions and make tracking more robust when the hand is rotated.

\subsection{Real-World Results}
We show real-world results obtained with our ISyHand setup for two of the four objects in Fig.~\ref{fig:realworldresults}.
By using tactile and penetration factors, our approach can correct for offsets in the visual pose estimate and yield a more geometrically plausible pose estimate of the object in the robot hand.
Please also refer to the supplementary material for videos and results for the other two objects (mustard bottle, tomato soup can).
Our implementation is real-time capable at approximately 13.3\,Hz on average on these sequences.

\section{Conclusion}

This paper proposes a novel approach for visuo-tactile object-pose estimation for multi-finger robot hands that are equipped with low-resolution tactile sensor pads at each inner surface of the hand.
We formulate pose estimation as a factor-graph optimization that includes a vision estimate, non-penetration constraints, and estimated contact points from the tactile sensor pads presently detecting contact.
We observe in simulation experiments that tactile information can improve accuracy and robustness over pure visual tracking in high-occlusion settings.
We also demonstrate our approach with a preliminary version of the ISyHand in a real-world setting and find that even low-resolution tactile information with simple contact thresholding yields poses that are geometrically better aligned with the hand. We will continue to improve our tactile sensor pads and develop more robust contact models in future work. 
A limitation of our tracking approach at present is that it assumes constant measurement variance. Outliers are handled by the robust cost function with a fixed hyperparameter. Since measurement uncertainty is not modeled in a data-dependent way, fast dynamic object motion and visual tracking failures are difficult to distinguish.
In future work, we plan to improve our tracker for dynamic object motion by incorporating a dynamics model or estimating the uncertainty of the visual pose estimate.

\addtolength{\textheight}{-2cm}

\section*{Acknowledgment}

The authors acknowledge support in preliminary studies by Mikel Zhobro and Umur Gogebakan.

\bibliographystyle{IEEEtranS}
\bibliography{refs.bib}

\begin{thebibliography}{10}
\providecommand{\url}[1]{#1}
\csname url@rmstyle\endcsname
\providecommand{\newblock}{\relax}
\providecommand{\bibinfo}[2]{#2}
\providecommand\BIBentrySTDinterwordspacing{\spaceskip=0pt\relax}
\providecommand\BIBentryALTinterwordstretchfactor{4}
\providecommand\BIBentryALTinterwordspacing{\spaceskip=\fontdimen2\font plus
\BIBentryALTinterwordstretchfactor\fontdimen3\font minus
  \fontdimen4\font\relax}
\providecommand\BIBforeignlanguage[2]{{%
\expandafter\ifx\csname l@#1\endcsname\relax
\typeout{** WARNING: IEEEtran.bst: No hyphenation pattern has been}%
\typeout{** loaded for the language `#1'. Using the pattern for}%
\typeout{** the default language instead.}%
\else
\language=\csname l@#1\endcsname
\fi
#2}}

\bibitem{allegrohand}
``Allegro hand,'' \hspace{-0.5pt}https://www.wonikrobotics.com/robot-hand.

\bibitem{shadowhand}
``Shadow hand,''
  \hspace{-0.5pt}https://www.shadowrobot.com/dexterous-hand-series/.

\bibitem{ahn2019_d435}
M.~S. Ahn, H.~Chae, D.~Noh, H.~Nam, and D.~W. Hong, ``Analysis and noise
  modeling of the {I}ntel {R}eal{S}ense {D435} for mobile robots,'' in
  \emph{Proc. of the International Conference on Ubiquitous Robots (UR)}.\hskip
  1em plus 0.5em minus 0.4em\relax {IEEE}, 2019, pp. 707--711.

\bibitem{anzai2020_vistac}
T.~Anzai and K.~Takahashi, ``Deep gated multi-modal learning: In-hand object
  pose changes estimation using tactile and image data,'' in \emph{Proc. of the
  {IEEE/RSJ} International Conference on Intelligent Robots and Systems
  (IROS)}, 2020, pp. 9361--9368.

\bibitem{bauza2023_tac2pose}
M.~Bauz{\'{a}}, A.~Bronars, and A.~Rodriguez, ``{Tac2Pose}: Tactile object pose
  estimation from the first touch,'' \emph{Int. J. Robotics Res.}, vol.~42,
  no.~13, pp. 1185--1209, 2023.

\bibitem{bimbo2012_vistac}
J.~Bimbo, S.~Rodr{\'{\i}}guez{-}Jim{\'{e}}nez, H.~Liu, X.~Song, N.~Burrus,
  L.~D. Seneviratne, M.~Abderrahim, and K.~Althoefer, ``Object pose estimation
  and tracking by fusing visual and tactile information,'' in \emph{Proc. of
  the International Conference on Multisensor Fusion and Integration for
  Intelligent Systems (MFI)}, 2012, pp. 65--70.

\bibitem{bishop2007_prml}
C.~M. Bishop, \emph{Pattern recognition and machine learning, 5th Edition},
  ser. Information science and statistics.\hskip 1em plus 0.5em minus
  0.4em\relax Springer, 2007.

\bibitem{Burns22-FRAI-Endowing}
R.~B. Burns, H.~Lee, H.~Seifi, R.~Faulkner, and K.~J. Kuchenbecker, ``Endowing
  a {NAO} robot with practical social-touch perception,'' \emph{Frontiers in
  Robotics and AI}, vol.~9, p. 840335, Apr. 2022.

\bibitem{caddeo2023_collawaretacpose}
G.~M. Caddeo, N.~A. Piga, F.~Bottarel, and L.~Natale, ``Collision-aware in-hand
  {6D} object pose estimation using multiple vision-based tactile sensors,'' in
  \emph{Proc. of the IEEE International Conference on Robotics and Automation
  (ICRA)}, 2023, pp. 719--725.

\bibitem{Calli17-IJRR-YCB}
B.~Calli, A.~Singh, J.~Bruce, A.~Walsman, K.~Konolige, S.~Srinivasa, P.~Abbeel,
  and A.~M. Dollar, ``{Y}ale-{CMU}-{B}erkeley dataset for robotic manipulation
  research,'' \emph{The International Journal of Robotics Research}, vol.~36,
  no.~3, pp. 261--268, 2017.

\bibitem{dellaert2017_factorgraphs}
F.~Dellaert and M.~Kaess, ``Factor graphs for robot perception,'' \emph{Found.
  Trends Robotics}, vol.~6, no. 1-2, pp. 1--139, 2017.

\bibitem{dikhale2022_vistac}
S.~Dikhale, K.~Patel, D.~Dhingra, I.~Naramura, A.~Hayashi, S.~Iba, and
  N.~Jamali, ``Visuotactile {6D} pose estimation of an in-hand object using
  vision and tactile sensor data,'' \emph{{IEEE} Robotics Autom. Lett.},
  vol.~7, no.~2, pp. 2148--2155, 2022.

\bibitem{du2021_robotgraspingreview}
G.~Du, K.~Wang, S.~Lian, and K.~Zhao, ``Vision-based robotic grasping from
  object localization, object pose estimation to grasp estimation for parallel
  grippers: a review,'' \emph{Artif. Intell. Rev.}, vol.~54, no.~3, pp.
  1677--1734, 2021.

\bibitem{handa2023_dextreme}
A.~Handa, A.~Allshire, V.~Makoviychuk, A.~Petrenko, R.~Singh, J.~Liu,
  D.~Makoviichuk, K.~V. Wyk, A.~Zhurkevich, B.~Sundaralingam, and Y.~S. Narang,
  ``{DeXtreme}: Transfer of agile in-hand manipulation from simulation to
  reality,'' in \emph{Proc. of the {IEEE} International Conference on Robotics
  and Automation (ICRA)}, 2023, pp. 5977--5984.

\bibitem{he2021_ffb6d}
Y.~He, H.~Huang, H.~Fan, Q.~Chen, and J.~Sun, ``{FFB6D}: A full flow
  bidirectional fusion network for {6D} pose estimation,'' in \emph{Proc. of
  the {IEEE}/{CVF} Conference on Computer Vision and Pattern Recognition
  ({CVPR})}, June 2021.

\bibitem{hinterstoisser2012_tless}
S.~Hinterstoisser, V.~Lepetit, S.~Ilic, S.~Holzer, G.~R. Bradski, K.~Konolige,
  and N.~Navab, ``Model based training, detection and pose estimation of
  texture-less {3D} objects in heavily cluttered scenes,'' in \emph{Proc. of
  the Asian Conference on Computer Vision (ACCV)}, 2012.

\bibitem{izatt2017_vistac}
G.~Izatt, G.~Mirano, E.~H. Adelson, and R.~Tedrake, ``Tracking objects with
  point clouds from vision and touch,'' in \emph{Proc. of the IEEE
  International Conference on Robotics and Automation (ICRA)}, 2017, pp.
  4000--4007.

\bibitem{kelestemur2022_tacpose}
T.~Kelestemur, R.~Platt, and T.~Padir, ``Tactile pose estimation and policy
  learning for unknown object manipulation,'' in \emph{Proc. of the Int. Conf.
  on Autonomous Agents and Multiagent Systems (AAMAS)}, 2022, pp. 742--750.

\bibitem{koval2025_tacposeplanarpush}
M.~C. Koval, N.~S. Pollard, and S.~S. Srinivasa, ``Pose estimation for planar
  contact manipulation with manifold particle filters,'' \emph{Int. J. Robotics
  Res.}, vol.~34, no.~7, pp. 922--945, 2015.

\bibitem{li2018_deepim}
Y.~Li, G.~Wang, X.~Ji, Y.~Xiang, and D.~Fox, ``{DeepIM}: Deep iterative
  matching for {6D} pose estimation,'' \emph{International Journal of Computer
  Vision}, vol. 128, no.~3, pp. 657--678, Nov. 2019.

\bibitem{liang2020_inhandposeparallelsim}
J.~Liang, A.~Handa, K.~V. Wyk, V.~Makoviychuk, O.~Kroemer, and D.~Fox,
  ``In-hand object pose tracking via contact feedback and {GPU}-accelerated
  robotic simulation,'' in \emph{Proc. of the IEEE International Conference on
  Robotics and Automation (ICRA)}, 2020, pp. 6203--6209.

\bibitem{lin2023_tactileekf}
Q.~Lin, C.~Yan, Q.~Li, Y.~Ling, Y.~Zheng, W.~Lee, Z.~Wan, B.~Huang, and X.~Liu,
  ``Tactile-based object pose estimation employing extended kalman filter,'' in
  \emph{Proc. of the International Conference on Advanced Robotics and
  Mechatronics (ICARM)}.\hskip 1em plus 0.5em minus 0.4em\relax {IEEE}, 2023,
  pp. 118--123.

\bibitem{lipson2022_coupled}
L.~Lipson, Z.~Teed, A.~Goyal, and J.~Deng, ``Coupled iterative refinement for
  {6D} multi-object pose estimation,'' in \emph{Proc. of the {IEEE}/{CVF}
  Conference on Computer Vision and Pattern Recognition ({CVPR})}.\hskip 1em
  plus 0.5em minus 0.4em\relax {IEEE}, June 2022.

\bibitem{theseus}
L.~Pineda, T.~Fan, M.~Monge, S.~Venkataraman, P.~Sodhi, R.~T.~Q. Chen,
  J.~Ortiz, D.~DeTone, A.~S. Wang, S.~Anderson, J.~Dong, B.~Amos, and
  M.~Mukadam, ``Theseus: {A} library for differentiable nonlinear
  optimization,'' in \emph{Proc. of Advances in Neural Information Processing
  Systems (NeurIPS)}, 2022.

\bibitem{rostel2022_tacestimator}
L.~R{\"{o}}stel, L.~Sievers, J.~Pitz, and B.~B{\"{a}}uml, ``Learning a state
  estimator for tactile in-hand manipulation,'' in \emph{Proc. of the
  {IEEE/RSJ} International Conference on Intelligent Robots and Systems
  (IROS)}, 2022, pp. 4749--4756.

\bibitem{shaw2023leaphand}
K.~Shaw, A.~Agarwal, and D.~Pathak, ``Leap hand: Low-cost, efficient, and
  anthropomorphic hand for robot learning,'' \emph{Proc. of Robotics: Science
  and Systems (RSS)}, 2023.

\bibitem{sodhi2022_patchgraph}
P.~Sodhi, M.~Kaess, M.~Mukadam, and S.~Anderson, ``{PatchGraph}: In-hand
  tactile tracking with learned surface normals,'' in \emph{Proc. of the IEEE
  International Conference on Robotics and Automation (ICRA)}, 2022, pp.
  2164--2170.

\bibitem{strecke2021_diffsdfim}
M.~Strecke and J.~Stueckler, ``{DiffSDFSim}: Differentiable rigid-body dynamics
  with implicit shapes,'' in \emph{Proc. of the International Conference on 3D
  Vision (3DV)}, 2021, pp. 96--105.

\bibitem{todorov2012_mujoco}
E.~Todorov, T.~Erez, and Y.~Tassa, ``{MuJoCo}: {A} physics engine for
  model-based control,'' in \emph{International Conference on IEEE/RSJ
  Intelligent Robots and Systems (IROS)}.\hskip 1em plus 0.5em minus
  0.4em\relax {IEEE}, 2012, pp. 5026--5033.

\bibitem{wan2024_vint6d}
\BIBentryALTinterwordspacing
Z.~Wan, Y.~Ling, S.~Yi, L.~Qi, W.~W. Lee, M.~Lu, S.~Yang, X.~Teng, P.~Lu,
  X.~Yang, M.-H. Yang, and H.~Cheng, ``{V}in{T}-6{D}: A large-scale
  object-in-hand dataset from vision, touch and proprioception,'' in
  \emph{Proceedings of the International Conference on Machine Learning
  (ICML)}, 2024, pp. 49\,921--49\,940. [Online]. Available:
  \url{https://proceedings.mlr.press/v235/wan24d.html}
\BIBentrySTDinterwordspacing

\bibitem{wang2021_gdr}
G.~Wang, F.~Manhardt, F.~Tombari, and X.~Ji, ``{GDR-Net}: Geometry-guided
  direct regression network for monocular {6D} object pose estimation,'' in
  \emph{Proc. of the IEEE/CVF Conference on Computer Vision and Pattern
  Recognition (CVPR)}, June 2021.

\bibitem{wang2019_nocs}
H.~Wang, S.~Sridhar, J.~Huang, J.~Valentin, S.~Song, and L.~J. Guibas,
  ``Normalized object coordinate space for category-level {6D} object pose and
  size estimation,'' in \emph{Proc. of the IEEE/CVF Conference on Computer
  Vision and Pattern Recognition (CVPR)}, 2019, pp. 2642--2651.

\bibitem{wen2023_foundationpose}
B.~Wen, W.~Yang, J.~Kautz, and S.~Birchfield, ``{FoundationPose}: Unified {6D}
  pose estimation and tracking of novel objects,'' in \emph{Proc. of IEEE/CVF
  Conference on Computer Vision and Pattern Recognition (CVPR)}, 2024, to
  appear, preprint abs/2312.08344.

\bibitem{xiang2018_posecnn}
Y.~Xiang, T.~Schmidt, V.~Narayanan, and D.~Fox, ``{PoseCNN}: A convolutional
  neural network for {6D} object pose estimation in cluttered scenes,'' in
  \emph{Proc. of Robotics: Science and Systems {XIV}}, June 2018.

\bibitem{xu2023dexterous}
K.~Xu, Z.~Hu, R.~Doshi, A.~Rovinsky, V.~Kumar, A.~Gupta, and S.~Levine,
  ``Dexterous manipulation from images: Autonomous real-world {RL} via substep
  guidance,'' in \emph{Proc. of the IEEE International Conference on Robotics
  and Automation (ICRA)}.\hskip 1em plus 0.5em minus 0.4em\relax IEEE, 2023,
  pp. 5938--5945.

\bibitem{zhao2023_fingerslam}
J.~Zhao, M.~Bauz{\'{a}}, and E.~H. Adelson, ``{FingerSLAM}: Closed-loop unknown
  object localization and reconstruction from visuo-tactile feedback,'' in
  \emph{Proc. of the {IEEE} International Conference on Robotics and Automation
  (ICRA)}, 2023, pp. 8033--8039.

\end{thebibliography}

\end{document}